\documentclass[11pt,a4paper]{article}
\usepackage[hyperref]{acl2020}
\usepackage{times}
\usepackage{latexsym}

\usepackage{microtype}
\usepackage{subfigure}
\usepackage{booktabs} 
\usepackage{wrapfig}
\usepackage{tcolorbox}

\usepackage[utf8]{inputenc} 
\usepackage[T1]{fontenc}    
\usepackage{amsfonts}       
\usepackage{nicefrac}       
\usepackage{subfigure}
\usepackage{xcolor}
\usepackage{adjustbox}
\usepackage{bm}
\usepackage{amsmath}
\usepackage{amssymb}
\usepackage{algorithm}
\usepackage{algorithmic}
\usepackage{multirow}
\usepackage{adjustbox}

\aclfinalcopy

\newcommand{\RN}[1]{%
	\textup{\lowercase\expandafter{\it \romannumeral#1}}%
}

\usepackage{algorithm}
\usepackage{algorithmic}

\usepackage{amsmath}
\usepackage{amsfonts}
\usepackage{amssymb}

\newcommand{\beq}{\vspace{0mm}\begin{equation}}
\newcommand{\eeq}{\vspace{0mm}\end{equation}}
\newcommand{\beqs}{\vspace{0mm}\begin{eqnarray}}
\newcommand{\eeqs}{\vspace{0mm}\end{eqnarray}}
\newcommand{\barr}{\begin{array}}
\newcommand{\earr}{\end{array}}
\newcommand{\av}{{\boldsymbol a}}

\newcommand{\fv}{{\boldsymbol f}}

\newcommand{\sv}{{\boldsymbol s}}

\newcommand{\wv}{{\boldsymbol w}}

\newcommand{\Ov}{{\boldsymbol O}}

\newcommand{\Xv}{{\boldsymbol X}}

\usepackage[bottom]{footmisc}

\def\aclpaperid{152}

\begin{document}

\title{Improving Adversarial Text Generation by Modeling the Distant Future}

\author{Ruiyi Zhang$^{1}$, Changyou Chen$^{2}$, Zhe Gan$^{3}$, Wenlin Wang$^{4}$ \\
	\textbf{Dinghan Shen}$^{3}$, \textbf{Guoyin Wang}$^{1}$,  \textbf{Zheng Wen}$^{5}$, \textbf{Lawrence Carin}$^{1}$
	\\
	$^{1}$ Duke University \quad$^{2}$ University at Buffalo \quad$^{3}$ Microsoft Dynamics 365 AI\\ \quad$^{4}$ Citadel LLC \quad$^{5}$ DeepMind\\
	{\tt ryzhang@cs.duke.edu } \\
}

\maketitle
\begin{abstract}
Auto-regressive text generation models usually focus on local fluency, and may cause inconsistent semantic meaning in long text generation. Further, automatically generating words with similar semantics is challenging, and hand-crafted linguistic rules are difficult to apply. We consider a \textit{text planning} scheme and present a model-based imitation-learning approach to alleviate the aforementioned issues. Specifically, we propose a novel guider network to focus on the generative process over a longer horizon, which can assist next-word prediction and provide intermediate rewards for generator optimization. Extensive experiments demonstrate that the proposed method leads to improved performance. 
\end{abstract}

\section{Introduction}
Text generation is an important area of investigation within machine learning. Recent work has shown excellent performance on a number of tasks, by combining reinforcement learning (RL) and generative models. Example applications include image captioning \cite{ren2017deep,rennie2016self}, text summarization \cite{li2018actor,paulus2017deep,rush2015neural}, and adversarial text generation \cite{guo2017long,lin2017adversarial,yu2017seqgan,zhang2017adversarial,zhu2018texygen}. The sequence-to-sequence framework (Seq2Seq) \cite{sutskever2014sequence} is a popular technique for text generation. However, models from such a setup are typically trained to predict the next token given previous ground-truth tokens as input, causing what is termed exposure bias \cite{ranzato2015sequence}. 
By contrast, sequence-level training with RL provides an effective means of solving this challenge, by treating text generation as a sequential decision-making problem. By directly optimizing an evaluation score (cumulative rewards) \cite{ranzato2015sequence}, state-of-the-art results have been obtained in many text-generation tasks \cite{paulus2017deep,rennie2016self}. 
However, one problem in such a framework is that rewards in RL training are particularly sparse, since a scalar reward is typically only available after an entire sequence has been generated. Furthermore, the recurrent models focus more on local fluency, and may cause inconsistent semantic meanings for long text generation. 

For RL-based text generation, most existing works rely on a model-free framework, which has been criticized for its high variance and poor sample efficiency \cite{sutton1998reinforcement}.
On the other hand, while model-based RL methods do not suffer from these issues, they are usually difficult to train in complex environments. Further, a learned policy is usually restricted by the capacity of an environment model. Recent developments on model-based RL \cite{gu2016continuous,kurutach2018model,nagabandi2017neural} combine the advantages of these two approaches, and have achieved improved performance by learning a model-free policy, assisted by an environment model. In addition, model-based RL has been employed recently to solve problems with extremely sparse rewards, with curiosity-driven methods \cite{pathak2017curiosity}.

In this paper, we propose a \textit{model-based imitation-learning} method to overcome the aforementioned issues in text-generation tasks. Our main idea is to employ an explicit guider network to model the generation environment in the feature space of sentence tokens, used to emit intermediate rewards by matching the predicted features from the guider network and features from generated sentences. The guider network is trained to encode global structural information of training sentences, and thus is useful to guide next-token prediction in the generative process. Within the proposed framework, to assist the guider network, we also develop a new type of self-attention mechanism to provide high-level planning-ahead information and maintain consistent semantic meaning. Our experimental results demonstrate the effectiveness of proposed methods.

\section{Background}
\paragraph{Text Generation Model}
Text generation models learn to generate a sentence $Y=(y_1, \ldots, y_T)$ of length $T$, possibly conditioned on some context $X$. Here each $y_t$ is a token from vocabulary $\mathcal{A}$. 
%
%
Starting from the initial state $\sv_0$, 
a recurrent neural network (RNN) produces a sequence of states $(\sv_1,\ldots,\sv_T)$ given an input sentence-feature representation $(e(y_1), \dots, e(y_T))$, where $e(\cdot)$ denotes a word embedding function mapping a token to its $d$-dimensional feature representation. 
The states are recursively updated with a function known as the {\it cell}: $\sv_t=h(\sv_{t-1}, e(y_t))$.
One typically assigns the following probability to an observation $y$ at location $t$: 
$p(y|Y_{<t}) = [\text{softmax}(g(\sv_{t}))]_y$. Together $(g,h)$ specifies a probabilistic model $\pi$, {\it i.e.}, 
\begin{align}
\log \pi(Y) = \sum_t \log p(y_t|Y_{<t}). 
\end{align}
\vspace{-4mm}

To train the model $\pi$, one typically uses maximum likelihood estimation (MLE), via minimizing the cross-entropy loss, \textit{i.e.}, ${J}_{\text{MLE}}(\pi)=-\mathbb{E}[\log \pi(Y)]$.
In order to generate sentence $Y^s$ from a (trained) model, one iteratively applies the following operations:
\begin{align}
{y}^s_{t+1} &\sim \text{Multi}(1, \text{softmax}(g(\sv_{t}))),\\
\sv_t &= h(\sv_{t - 1}, e({y}^s_t))\,,
\end{align}
where $\text{Multi}(1,\cdot)$ denotes one draw from a multinomial distribution.

\paragraph{Model-Based Imitation Learning}
Text generation can be considered as an RL problem with a large number of discrete actions, \textit{deterministic} transitions, and \textit{deterministic} terminal rewards. It can be formulated as a Markov decision process (MDP) $\mathcal{M} = \langle\mathcal{S}, \mathcal{A}, P, r, \gamma \rangle$, where 
$\mathcal{S}$ is the state space, $\mathcal{A}$ is the action space, $P$ is the deterministic
environment dynamics, $r(\sv,y)$ is a reward function, and $\gamma \in (0,1)$ is the discrete-time discount factor.
The policy $\pi_{\phi}$, parameterized by $\phi$, maps each state $\sv \in \mathcal{S}$ to a probability distribution over $\mathcal{A}$. The objective is to maximize the expected reward:
\vspace{-2mm}
\begin{equation}
\begin{aligned}\label{eq:policy_learning}
J(\pi) = \sum_{t=1}^{\infty}\mathbb{E}_{P, \pi}\left[\gamma^{t-1}\cdot r(\sv_t, y_t)\right].
\end{aligned}
\end{equation}
In model-based imitation learning \cite{baram2016model,cheng2019accelerating}, a model is built to make predictions for future state $\sv_{t+\triangle t}$ conditioned on the current state\footnote{~$\triangle t > 1$; the model predicts future states based on the collected trajectories.}, which can be used for action selection, {\it e.g.}, next-token generation. This model is typically a discrete-time system, taking the current state-action pair $(\sv_t, y_t)$ as input, and outputting an estimate of the future state $\sv_{t+\triangle t}$ at time $t+\triangle t$. At each step $t$, $y_{t}$ is chosen based on the model, and the model will re-plan with the updated information from the dynamics. This control scheme is different from a standard model-based method, and is referred to as \textit{model-predictive control} (MPC) \cite{nagabandi2017neural}. 
Note that in our setting, the state in RL typically corresponds to the current generated sentences $Y_{1,\ldots,t}$ instead of the RNN state of generator (decoder). 


\section{Proposed Model}
\begin{figure*}[t!] 
	\centering
	\includegraphics[width=1.0\linewidth]{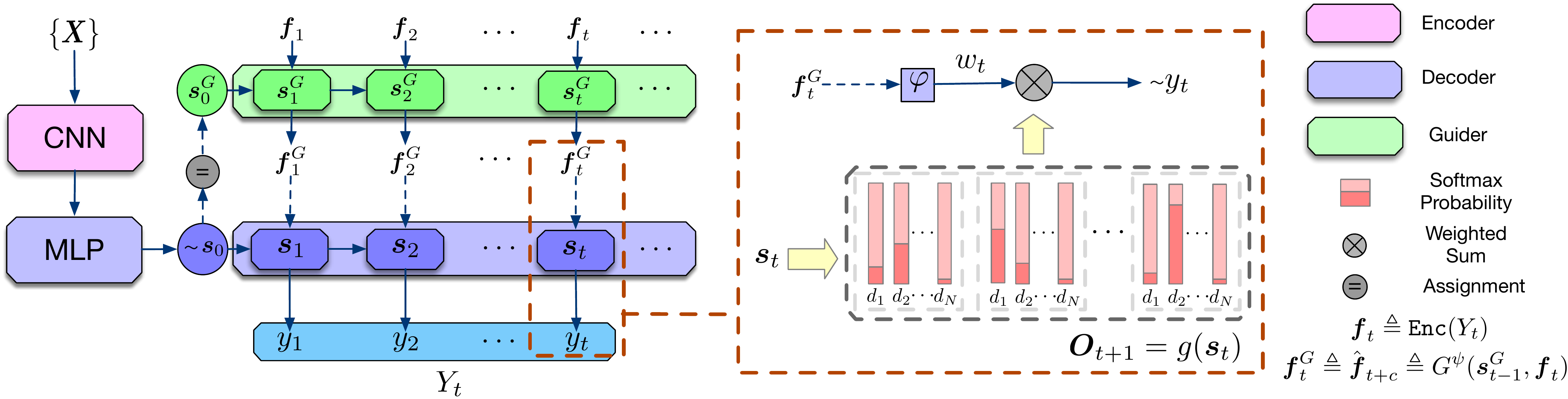}  
	\caption{Model overview of text generation with a guider network. Solid lines mean gradients are backpropagated in training; dash lines mean gradients are not backpropagated. CNN is the feature extractor, and MLP outputs the parameters of the Gaussian density which is compatible with the initial state of the LSTM Guider and Decoder.}
	\label{fig:Model}
	\vspace{-2mm}
\end{figure*}

The model is illustrated in Figure~\ref{fig:Model}, with an autoeocoder (AE) structure for sentence feature extraction and generation. The encoder is shared for sentences from both training data and generated data, as explained in detail below. Overall, text generation can be formulated as an imitation-learning problem.
At each timestep $t$, the agent, also called a generator (which corresponds to the LSTM decoder), takes the current LSTM state as input, denoted as $\sv_{t}$. The policy $\pi_{\phi}(\cdot|\sv_t)$ parameterized by $\phi$ is a conditional generator, to generate the next token (action) given $\sv_t$, the \textit{observation} representing the current generated sentence. 
The objective of text generation is to maximize the total reward as in \eqref{eq:policy_learning}. We detail the components for our proposed model in the following subsections.
%
%

\subsection{The Guider Network}
The guider network, implemented as an RNN with LSTM units, is adopted to model \textit{environment dynamics} to assist text generation. The idea is to train a guider network such that its predicted sentence features at each time step are used to assist next-word generation and construct intermediate rewards, which in turn are used to optimize the sentence generator. Denote the guider network as $G^{\psi}(\sv_{t-1}^G, \fv_t)$, with parameters $\psi$ and input arguments $(\sv_{t-1}^G, \fv_t)$ at time $t$, to explicitly write out the dependency on the {\em guider network} latent state $\sv_{t-1}^G$ from the previous time step. Here $\fv_t$ is the input to the LSTM guider, which represents the feature of the current generated sentence extracted by an encoder network. Specifically, let the current generated sentence be $Y_{1\ldots t}$ (encouraged to be the same as parts of a training sentence in training), with $\fv_t$ calculated as:
$\fv_t = \mbox{Enc}(Y_{1\ldots t})$.
%
%
The initial state of the guider network is the encoded feature of a true input sentence by the same convolutional neural network (CNN), {\it i.e.}, $\sv_0^G = \mbox{Enc}(\Xv)$, where $\mbox{Enc}(\cdot)$ denotes the encoder transformation, implemented with a CNN~\cite{zhang2017adversarial}.
Importantly, the input to the guider network, at each time point, is defined by features from the entire sentence generated to that point. This provides an important ``guide'' to the 
LSTM decoder, accounting for the global properties of the generated text.
\vspace{-4mm}
\paragraph{Text Generation with Planning} We first explain how one uses the guider network to guide next-word generation for the generator (the LSTM decoder in Figure~\ref{fig:Model}). Our framework is inspired by the MPC method \cite{nagabandi2017neural}, and can be regarded as a type of plan-ahead attention mechanism. Given the feature $\fv_{t}$ at time $t$ from the current input sentence, the guider network produces a prediction $G^{\psi}(\sv^G_{t-1}, \fv_{t})$ as a future feature representation, by feeding $\fv_{t}$ into the LSTM guider. Since the training of the guider network is based on real data (detailed in the next paragraph), the predicted feature contains global-structure information of the training sentences. To utilize such information to predict the next word, we combine the predicted feature with the output of the decoder by constructing an attention-like mechanism. Specifically, we first apply a linear transformation $\varphi$ on the predicted feature $G^{\psi}(\sv^G_{t-1}, \fv_{t})$, forming a weight vector $\wv_t \triangleq \varphi\left(G^{\psi}(\sv^G_{t-1}, \fv_{t})\right)$. The weight $\wv_t$ is applied to the output $\Ov_t$ of the LSTM decoder by an element-wise multiplication operation. The result is then fed into a softmax layer to generate the next token $y_{t}$. Formally, the generative process is written as:
\vspace{-2mm}
\begin{align}
\Ov_{t} &= g(\sv_{t-1}), ~~
\wv_t = \varphi(G^{\psi}(\sv^G_{t-1}, \fv_{t})),\\
y_{t}&\sim\text{Multi}(1,\text{softmax}(\Ov_{t}\cdot\wv_t)), \\
\sv^G_{t} &= h^G(\sv^G_{t-1}, \fv_{t}),~~~~
\sv_{t} = h(\sv_{t-1}, e(y_{t}))~.
\end{align}

\paragraph{Guider Network Training} 
Given a sentence of feature representations $(\fv_1$, $\fv_2$, \ldots $\fv_T)$ for a training sentence, we seek to update the guider network such that it is able to predict $\fv_{t+c}$ given $\fv_{t}$, where $c > 0$ is the number of steps that are looked ahead. We implement this by forcing the predicted feature, $G^{\psi}(\sv_{t}^G, \fv_{t})$, to match both the sentence feature $\fv_{t+c}$ (first term in \eqref{eq:JG}) and the corresponding feature-changing direction (second term in \eqref{eq:JG}). This is formalized by maximizing an objective function of the following form at time $t$:
\begin{align}\label{eq:JG}
J^\psi_G&= \mathcal{D}_{\cos}\left(\fv_{t+c},~ G^{\psi}(\sv^G_{t-1}, \fv_{t})\right)\\ &+\mathcal{D}_{\cos}\left(\fv_{t+c}-\fv_{t},~ G^{\psi}(\sv_{t-1}^G, \fv_{t})-\fv_{t}\right)~,\nonumber
\end{align}
where $\mathcal{D}_{\cos}(\cdot, \cdot)$ denotes the cosine similarity\footnote{We found that the cosine similarity worked better than the $l_2$-norm.}. By maximizing \eqref{eq:JG}, an ideal guider network should be able to predict the true next words conditioned on the current word in a sentence. As a result, the prediction is used to construct an intermediate reward, used to update the generator (the LSTM decoder), as described further below. 
\subsection{Feature-Matching Rewards and Generator Optimization}
As in many RL-based text-generation methods, such as SeqGAN \cite{yu2017seqgan} and LeakGAN \cite{guo2017long}, the generator is updated based on policy-gradient methods. As a result, collecting rewards in the generation process is critical. Though SeqGAN~\cite{yu2017seqgan} has proposed to use rollout to get rewards for each generated word, the variance of the rewards is typically too high to be useful practically. In addition, the computational cost may be too high for practical use. We below describe how to use the proposed guider network to define intermediate rewards, leading to a definition of feature-matching reward.

\paragraph{Feature-Matching Rewards}
We first define an intermediate reward to generate a particular word. The idea is to match the ground-truth features from the CNN encoder in Figure~\ref{fig:Model} with those generated from the guider network. Equation \eqref{eq:JG} indicates that the further the generated feature is from the true feature, the smaller the reward should be. To this end, for each time $t$, we define the intermediate reward for generating the current word as:
\begin{equation}
\begin{aligned} 
r^g_t = \dfrac{1}{2c} \sum_{i=1}^{c} (&\mathcal{D}_{cos} (\fv_t, \hat{\fv}_t) + \\
 &\mathcal{D}_{cos} (\fv_t-\fv_{t-i}, \hat{\fv}_t-\fv_{t-i}))~, \nonumber
\end{aligned}
\end{equation}
where $\hat{\fv}_t = G^{\psi}(s_{t-c-1}^G, \fv_{t-c})$ is the predicted feature.
Intuitively, $\fv_t-\fv_{t-i}$ measures the difference between the generated sentences in feature space; the reward is high if it matches the predicted feature transition $\hat{\fv}_t-\fv_{t-i}$ from the guider network. At the last step of text generation, {\it i.e.}, $t = T$, the corresponding reward measures the quality of the whole generated sentence, thus it is called a final reward. The final reward is defined differently from the intermediate reward, discussed below for both the unconditional- and conditional-generation cases.
%

Note that a token generated at time $t$ will influence not only the rewards received at that time but also the rewards at subsequent time steps. Thus we propose to define the cumulative reward, $\sum_{i=t}^{T} \gamma^i r^g_i$ with $\gamma$ a discount factor, as a {\em feature-matching reward}. Intuitively, this encourages the generator to focus on achieving higher long-term rewards.
Finally, in order to apply policy gradient to update the generator, we combine the feature-matching reward with the problem-specific final reward, to form a $Q$-value reward specified below. 

Similar to SeqGAN, the final reward is defined as the output of a discriminator, evaluating the quality of the whole generated sentence, {\it i.e.}, the smaller the output, the less likely the generation is a true sentence. As a result, we combine the adversarial reward $r^f\in[0,1]$ by the discriminator~\cite{yu2017seqgan} with the guider-matching rewards, to define a $Q$-value reward as $Q_t  = (\sum_{i=t}^{T} \gamma^i r^g_{i}) \times r^f$.
\paragraph{Generator Optimization}
The generator is initialized by pre-training on sentences with an autoencoder structure, based on MLE training. After that, the final $Q$-value reward $Q_t$ is used as a reward for each time $t$, with standard policy gradient optimization methods to update the generator. Specifically, the policy gradient is
\begin{equation}
\begin{aligned}\nonumber
\nabla_{\phi}J &=\mathbb{E}_{(\sv_{t-1}, y_t)\sim \rho_{\pi}}\left[ Q_t \nabla_{\phi} \log p(y_t| \sv_{t-1}; \phi, \varphi)\right]\,,\\
\nabla_{\varphi}J &=\mathbb{E}_{(\sv_{t-1}, y_t)\sim \rho_{\pi}}\left[ Q_t \nabla_{\varphi} \log p(y_t| \sv_{t-1}; \phi, \varphi)\right] \,,
\end{aligned}
\end{equation}
where $p(y_t| \sv_{t-1}; \phi,\varphi)$ is the probability of generating $y_t$ given $\sv_{t - 1}$ in the generator. Algorithm 1 describes the proposed model-based imitation learning framework for text generation.
\begin{algorithm}[t!]
	\caption{Model-based Imitation Learning for Text Generation}
	\label{algo:algo1}
	\begin{algorithmic}[1]
		\REQUIRE generator policy $\pi^{\phi}$; guider network $G^\psi$; a sequence dataset $\{X_{1 \ldots T}\}$ by some expert policy.
		\STATE Initialize $G^\psi$, $D^\theta$ with random weights.
		\WHILE{Imitation Learning phase} 
		\STATE Update generator $\pi^{\phi}$, guider $G^\psi$ with MLE loss.
		\ENDWHILE
		\WHILE{Reinforcement Learning phase}
		\STATE{Generate a sequence $Y_{1\ldots T} \sim \pi^{\phi}$.}
		\STATE {Compute $Q_t$, and update $\pi^\phi$.}
		\ENDWHILE
	\end{algorithmic}
\end{algorithm}

\paragraph{Model-based or Model-free}
Text generation seeks to generate the next word (action) given the current (sub-)sentence (state). 
The generator is considered as an agent that learns a policy to predict the next word given its current state. 
In previous work~\cite{ranzato2015sequence}, a metric reward is given and the generator is trained to only maximize the metric reward by trial, thus this is model-free learning. In the proposed method, the guider network models the environment dynamics, and is trained by minimizing the cosine similarity between the prediction and the ground truth on real text. For generator training, it maximizes the reward which is determined by the metric and guider network, and thus is model-free learning with model-based boosting~\cite{gu2016continuous}. 
The model predictive control scheme is included in our method, where the guider network is used to help next-word selection at each time-step. 
\section{Extension to Non-parallel Text Style Transfer}

\begin{figure}[t!] 
	\centering
	\hspace{-3mm}
	\includegraphics[width=1.01\linewidth]{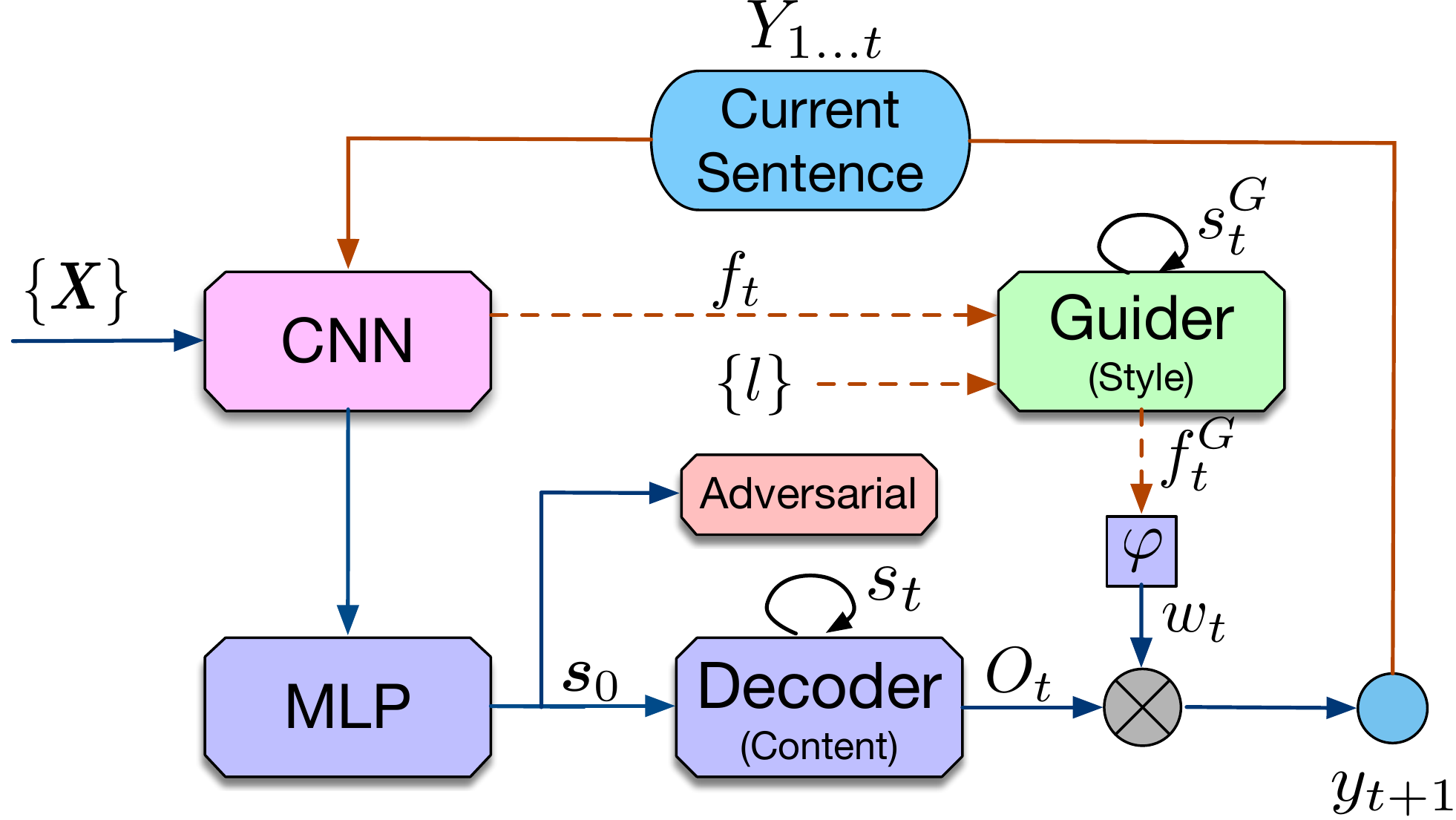}  
	\vspace{-2mm}
	\caption{Guided style transfer: the Guider network controls the sentiment in the higher level, and the Generator focuses on preserving content in the lower level. }
	\label{fig:Model2}
	\vspace{-2mm}
\end{figure}

As illustrated in Figure \ref{fig:Model2}, our framework naturally provides a way for style transfer, where the guider network plays the role of style selection, and the generator only focuses on maintaining content without considering the styles. 
To make the guider network focus on the guidance of styles, we assign the label $l$ as the initial state $\sv^G_0$ of the guider network. Specifically, at each step $t$, we feed the current sentence representation $\fv_t$ and label $l$ into the guider network:
\begin{align}
\Ov_{t} &= g(\sv_{t-1}), ~~
\wv_t = \varphi(G^{\psi}(\sv^G_{t-1}, [\fv_{t}, l])),\\
y_{t}&\sim\text{Multi}(1,\text{softmax}(\Ov_{t}\cdot\wv_t)).
\end{align}

For the generator, we put an adversarial regularizer on the encoded latent $\sv_0(X)$ and penalize it if it contains the sentiment information, by maximizing the entropy, \textit{i.e.,} $\max \sum_{l} p(l|~\sv_0(X)) \log p(l|~\sv_0(X))$, where $p$ is a pre-trained classifier.
Intuitively, the generator gives candidate words represented by $O_t$, while the guider makes a choice implicitly by $w_t$ based on the sentiment information. The sentiment information is contained in $w_t$, while the content of the original sentence is represented by $O_t$. To achieve style-transfer, one feeds the original sentence $X$ with the target style label $l$ to get the transferred sentence $Y$ with style $l$.  Following previous work~\cite{hu2017controllable,yang2018unsupervised, cheng2020}, we adopt a classifier as the discriminator and the soft-argmax approach~\cite{kusner2016gans} for the update of generator instead of policy gradient~\cite{sutton1998reinforcement}.

\section{Related Work}

We first review related works that combine RL and GAN for text generation.
As one of the most representative models in this direction, 
SeqGAN~\cite{yu2017seqgan} adopts Monte-Carlo search to calculate rewards. However, such a method introduces high variance in policy optimization. There are a number of works proposed subsequently to improve the reward-generation process. For example, RankGAN~\cite{lin2017adversarial} proposes to replace the reward from the GAN discriminator with a ranking-based reward, MaliGAN \cite{che2017maximum} modifies the GAN objective and proposes techniques to reduce gradient variance, MaskGAN~\cite{fedus2018maskgan} uses a filling technique to define a $Q$-value reward for sentence completion,
RelGAN~\cite{nie2018relgan} uses a relational memory based generator for the long-distance dependency modeling, FMGAN~\cite{chen2018adversarial} uses a feature mover distance to match features of real and generated sentences inspired by optimal transport~\cite{chen2019improving,zhang2018policy}, 
and LeakGAN~\cite{guo2017long} tries to address the sparse-reward issue for long-text generation with hierarchical RL by utilizing the leaked information from a GAN discriminator. One problem of LeakGAN is that it tends to overfit the training data, yielding generated sentences that are often not diverse. By contrast, by relying on a model-based imitation learning approach, our method learns global-structure information, which generates more-diverse sentences, and can be extended to conditional text generation. \citet{zhang2020nested} designed a differentiable nested Wasserstein distance for semantic matching, which can be applied for further improvement.  

\begin{table*}[t!]
\centering
\begin{adjustbox}{scale=0.88, tabular=l rccc  rccc,center}
\toprule[1.2pt]
\textbf{Method} & \textbf{Test-BLEU-2} & \textbf{3} & \textbf{4} & \textbf{5} & \textbf{Self-BLEU-2} & \textbf{3} & \textbf{4}  \\
\midrule
SeqGAN \cite{yu2017seqgan}  & 0.820 & 0.604 & 0.361 & 0.211 & 0.807 & 0.577 & 0.278 \\
RankGAN \cite{lin2017adversarial} & 0.852 & 0.637 & 0.389 & 0.248 & 0.822 & 0.592 & 0.230 \\
GSGAN~\cite{kusner2016gans} & 0.810 & 0.566 & 0.335 & 0.197 & 0.785 & 0.522 & 0.230\\
TextGAN \cite{zhang2017adversarial} & 0.910 & 0.728 & 0.484 & 0.306 & 0.806 & 0.548 & 0.217  \\
LeakGAN \cite{guo2017long} & 0.922 & 0.797 & 0.602 & 0.416 & 0.912 & 0.825 & 0.689  \\
MLE~~\cite{caccia2018language}   & 0.902 & 0.706 & 0.470 & 0.392 & 0.787 &0.646 & 0.485 \\
\hline
GMGAN (ours) & 0.949 & 0.823 & 0.635 & 0.421 & 0.746 & 0.511 & 0.319\\
\bottomrule[1.2pt]
\end{adjustbox}
\vspace{-2mm}
\caption{Test-BLEU ($\uparrow$) and Self-BLEU ($\downarrow$) scores on Image COCO.}
\label{tab:cocoun}
\end{table*}

\begin{table*}[t!]
\centering
\begin{adjustbox}{scale=0.88,tabular=l rccc  rcc,center}
\toprule[1.2pt]
\textbf{Method} & \textbf{Test-BLEU-2} & \textbf{3} & \textbf{4} & \textbf{5} & \textbf{Self-BLEU-2} & \textbf{3} & \textbf{4}  \\
\midrule
SeqGAN \cite{yu2017seqgan}  & 0.630 & 0.354 & 0.164 & 0.087 & 0.728 & 0.411 & 0.139  \\
RankGAN \cite{lin2017adversarial} & 0.723 & 0.440 & 0.210 & 0.107 & 0.672 & 0.346 & 0.119 \\
GSGAN \cite{kusner2016gans} & 0.723 & 0.440 & 0.210 & 0.107 &  0.807 & 0.680 & 0.450\\
TextGAN \cite{zhang2017adversarial} & 0.777 & 0.529 & 0.305 & 0.161 & 0.806 & 0.662 & 0.448 \\
LeakGAN \cite{guo2017long} & 0.923 & 0.757 & 0.546 & 0.335 & 0.837 & 0.683 & 0.513 \\
MLE \cite{caccia2018language}& 0.902 &  0.706 & 0.470 & 0.392 &    0.787 & 0.646 & 0.485 \\
\hline
GMGAN (ours) & 0.923 & 0.727 & 0.491 & 0.303 & 0.814 & 0.576 & 0.328\\
\bottomrule[1.2pt]
\end{adjustbox}
\vspace{-2mm}
\caption{Test-BLEU ($\uparrow$) and Self-BLEU ($\downarrow$) scores on EMNLP2017 WMT News.}
\label{tab:wmtun}
\end{table*}
\vspace{-1mm}
RL techniques can also be used in other ways for text generation~\cite{bachman2015data}. For example, \citet{ranzato2015sequence} trained a Seq2Seq model by directly optimizing the BLEU/ROUGE scores with the REINFORCE algorithm. 
To reduce variance of the vanilla REINFORCE, \citet{bahdanau2016actor} adopted the actor-critic framework for sequence prediction. Furthermore, \citet{rennie2016self} trained a baseline algorithm with a greedy decoding scheme for the REINFORCE method. Note that all these methods can only obtain reward after a whole sentence is generated. Planning techniques in RL have also been explored to improve text generation \citep{gulcehre2017plan,serdyuk2018twin}. \citet{zhang2020nested} introduced the self-imitation scheme to exploit historical high-quality sentences for enhanced exploration. Compared to these related works, the proposed guider network can provide a planning mechanism and intermediate rewards. 

\section{Experiments}
We test the proposed framework on unconditional and conditional text generation tasks, and analyze the results to understand the performance gained by the guider network.
%
%
We also perform an ablation investigation on the improvements brought by each part of our proposed method, and consider non-parallel style transfer. All experiments are conducted on a single Tesla P100 GPU and implemented with TensorFlow and Theano. Details of the datasets, the experimental setup and
model architectures are provided in the Appendix. 

\subsection{Implementation Details}
\paragraph{Encoder as the feature extractor}
For unconditional generation, the feature extractor that generates inputs for the guider network shares the CNN part of the encoder. We stop gradients from the guider network to the encoder CNN in the training process. For conditional generation, we use a pre-trained feature extractor, trained similarly to the unconditional generation. 
\paragraph{Training procedure} 
As with many imitation-learning models~\cite{bahdanau2016actor,rennie2016self,sutskever2014sequence}, we first train the encoder-decoder part based on the off-policy data with an MLE loss. Then we use RL training to fine-tune the trained generator. We adaptively transfer the training from MLE loss to RL loss, similar to~\cite{paulus2017deep,ranzato2015sequence}.
\paragraph{Initial states}
We use the same initial state for both the generator and the guider networks. For conditional generation, the initial state is the encoded latent code of the conditional information for both training and testing. For unconditional generation, the initial state is the encoded latent code of a target sentence in training and random noise in testing.

\subsection{Adversarial Text Generation}

We focus on adversarial text generation, and compare our approach with a number of related works \cite{guo2017long,lin2017adversarial,yu2017seqgan,zhang2017adversarial,zhu2018texygen}. In this setting, a discriminator in the GAN framework is added to the model in Figure~\ref{fig:Model} to guide the generator to generate high-quality sentences. This is implemented by defining the final reward to be the output of the discriminator. All baseline experiments are implemented on the texygen platform~\cite{zhu2018texygen}. We adopt the BLEU score, referenced by the test set (test-BLEU, higher value implies better quality) and itself (self-BLEU, lower value implies better diversity)~\cite{zhu2018texygen} to evaluate quality of generated samples, where test-BLEU evaluates the reality of generated samples, and self-BLEU measures the diversity. A good generator should achieve both a high test-BLEU score and a low self-BLEU score. In practice, we use $\triangle t=c=4$ and $\gamma=0.25$. We call the proposed method guider-matching GAN (GMGAN) for unconditional text generation. More details of GMGAN are provided in Appendix D.
\paragraph{Short Text Generation: COCO Image Captions}
We use the COCO Image Captions Dataset, in which most sentences have a length of about 10 words. Since we consider unconditional text generation, only image captions are used as the training data. After preprocessing, we use 120,000 random sample sentences as the training set, and 10,000 as the test set. The BLEU scores with different methods are listed in Table~\ref{tab:cocoun}. We observe that
GMGAN performs significantly better than the baseline models. Specifically, besides achieving higher test-BLEU scores, the proposed method also generates samples with very good diversity in terms of self-BLEU scores. LeakGAN represents the state-of-the-art in adversarial text generation, however, its diversity measurement is relatively poor \cite{zhu2018texygen}. We suspect that the high BLEU score achieved by LeakGAN is due to its mode collapse on some good samples, resulting in high self-BLEU scores. Other baselines achieve lower self-BLEU scores since they cannot generate reasonable sentences.

\paragraph{Long Text Generation: EMNLP2017 WMT}
Following \cite{zhu2018texygen}, we use the News section in the EMNLP2017 WMT4 Dataset as our training data. 
The dataset consists of 646,459 words and 397,726 sentences. After preprocessing, the training
dataset contains 5,728 words and 278,686 sentences. The BLEU scores with different methods are provided in Table \ref{tab:wmtun}. Compared with other methods, LeakGAN and GMGAN achieve comparable test-BLEU scores, demonstrating high-quality generated sentences. Again, LeakGAN tends to over-fit on training data, leading to much higher (worse) self-BLEU scores. Our proposed GMGAN shows good diversity of long text generation with lower self-BLEU scores. Other baselines obtain both low self-BLEU and test-BLEU scores, leading to more random generations. 
\paragraph{Human Evaluation}
Simply relying on the above metrics is not sufficient to evaluate the proposed method~\cite{caccia2018language}. 
Following previous work~\cite{guo2017long}, we perform human evaluations using Amazon Mechnical Turk, evaluating the text quality based on readability and meaningfulness (whether sentences make sense) on the EMNNLP2017 WMT News dataset. We ask the worker to rate the input sentence with scores scaling from 1 to 5, with 1 as the worst score and 5 as the best. The detailed criteria is listed in Table~\ref{tab: rate_crit}.   We require all the workers to be native English speakers, with approval rate higher than 90\% and at least 100 assignments completed.

\begin{table}[t!]
	\label{tab: rate_crit}
	\small
	\centering
	\begin{adjustbox}{width=\linewidth,tabular=l p{6.3cm}, center}
		\toprule[1.2pt]
		\textbf{Scores} & ~~~~~~~~~~~~~~~~~~~~~~~~~~~~~~\textbf{Criteria}\\
		\midrule
        5 (Best)& It is consistent, informative, grammatically correct.\\
        4&It is grammatically correct and makes sense.\\
        3&It is mostly meaningful and with small grammatical error. \\
        2&It needs some time to understand and has grammatical errors.\\
        1 (Worst)& Meaningless, not readable.\\
		\bottomrule[1.2pt]
	\end{adjustbox} 
	\vspace{-2mm}
	\caption{Human evaluation rating criteria.}
	\label{tab: rate_crit}
\end{table}

We randomly sample 100 sentences generated by each model. Ten native English speakers on Amazon Mechanical Turk are asked to rate each sentence. The average human rating scores are shown in Table~\ref{tab:humanscore}, indicating GMGAN achieves higher human scores compared to other methods. 
As examples, Table \ref{tab: WMTexample} illustrates some generated samples by
GMGAN and its baselines. The performance on the two datasets indicates that the generated sentences of GMGAN are of higher global consistency and better readability than SeqGAN and LeakGAN. More generated examples are provided in the Appendix.

\begin{table}[t!]
	\begin{adjustbox}{width=\linewidth,tabular=l c  c  c c, center}
		\toprule[1.2pt]
		\textbf{Methods} & \textbf{MLE} & \textbf{SeqGAN} & \textbf{RankGAN} & \textbf{GSGAN} \\
		\midrule
		\textbf{Human scores} & 2.45\small{$\pm$0.14} & 2.57\small{$\pm$0.15} & 2.91\small{$\pm$0.17}  & 2.48\small{$\pm$0.14} \\
		\midrule
		\textbf{Methods} & \textbf{textGAN} & \textbf{LeakGAN} & \textbf{GMGAN} & \textbf{Real} \\
		\midrule
		\textbf{Human scores} & 3.11\small{$\pm$0.16} & 3.47\small{$\pm$0.15} & 3.89\small{$\pm$0.15}  & 4.21\small{$\pm$0.14} \\
		\bottomrule[1.2pt] 
	\end{adjustbox}
	\vspace{-2mm}
	\caption{
		Results of human evaluation with different methods on EMNLP2017 WMT dataset.
	}
	\label{tab:humanscore}
\end{table}

\begin{table*}[t!]
	\centering
	\begin{tcolorbox}
		\begin{adjustbox}{scale=0.8,tabular= l p{5.5cm} p{11cm},center}
			\textbf{Method} & \textbf{COCO Image Captions} & \textbf{EMNLP2017 WMT News}\\ 
			\midrule
			\textbf{SeqGAN}& (1) A person and black wooden table. \newline
			(2) A closeup of a window at night. & (1) She added on a page where it was made clear more old but public got said.\newline
			(2) I think she're guys in four years , and more after it played well enough. 
			\\ \hline
			\textbf{LeakGAN}&(1) A bathroom with a black sink and a white toilet next to a tub. \newline 
			(2) A man throws a Frisbee across the grass covered yard.&(1)"I'm a fan of all the game, I think if that's something that I've not," she said, adding that he would not be decided. \newline (2) The UK is Google' s largest non-US market, he has added "20, before the best team is amount of fewer than one or the closest home or two years ago.\\ \hline
			\textbf{GMGAN}& (1) Bicycles are parked near a row of large trees near a sidewalk. \newline
			(2) A married couple posing in front of a piece of birthday cake. & (1) "Sometimes decisions are big, but they're easy to make," he told The Sunday Times in the New Year.\newline (2) A BBC star has been questioned by police on suspicion of sexual assault against a 23-year-old man , it was reported last night.
		\end{adjustbox}
		\vspace{-2mm}
	\end{tcolorbox}
	\vspace{-2mm}
	\caption{ 
		Examples of generated samples with different methods on COCO and EMNLP datasets.
	}
	\label{tab: WMTexample}
\end{table*}

\paragraph{Ablation Study}
We conduct ablation studies on long text generation to investigate the improvements brought by each part of our proposed method. We first test the benefits of using the guider network. Among the methods compared, Guider is the standard MLE model with the guider network. We further compare RL training with $\RN{1})$ only final rewards
, $\RN{2})$ only feature-matching rewards, and $\RN{3})$ combining both rewards, namely GMGAN. The results are shown in Table~\ref{tab:ablation}. We observe that guider network plays an important role in improving the performance. RL training with final rewards given by a discriminator typically damages the generation quality, but feature-matching reward produces sentences with much better diversity due to the ability of exploration. 

\begin{table}[t!]
	\centering
	\begin{adjustbox}{width=\linewidth,tabular=l c  c  c  c c  , center}
		\toprule[1.2pt]
		\textbf{Methods} & \textbf{MLE} & \textbf{Guider} & \textbf{Final} & \textbf{Stepwise} &  \textbf{GMGAN}\\
		\midrule
		Test-BLEU-2  &0.761& 0.920 & 0.843 &  0.914  & 0.923\\
		~~~~~~~~BLEU-3  &0.468& 0.723 & 0.623 &  0.704  & 0.727\\
		~~~~~~~~BLEU-4  &0.230& 0.489 & 0.390 &  0.457  & 0.491\\
		~~~~~~~~BLEU-5  &0.116& 0.289 & 0.221 &  0.276  & 0.303\\
		\midrule
		Self-BLEU-2  &0.664& 0.812 & 0.778 &  0.798  & 0.814\\
		~~~~~~~~BLEU-3 &0.338& 0.589 & 0.525 &  0.563  & 0.576\\
		~~~~~~~~BLEU-4  &0.113& 0.360 & 0.273 &  0.331  & 0.328\\
		\bottomrule[1.2pt]
	\end{adjustbox}
	\vspace{-2mm}
	\caption{Ablation study on EMNLP2017 WMT.
	}
	\vspace{-2mm}
	\label{tab:ablation}
\end{table}

\begin{figure}[t!] \centering\hspace{-2mm}
	\includegraphics[width=\linewidth]{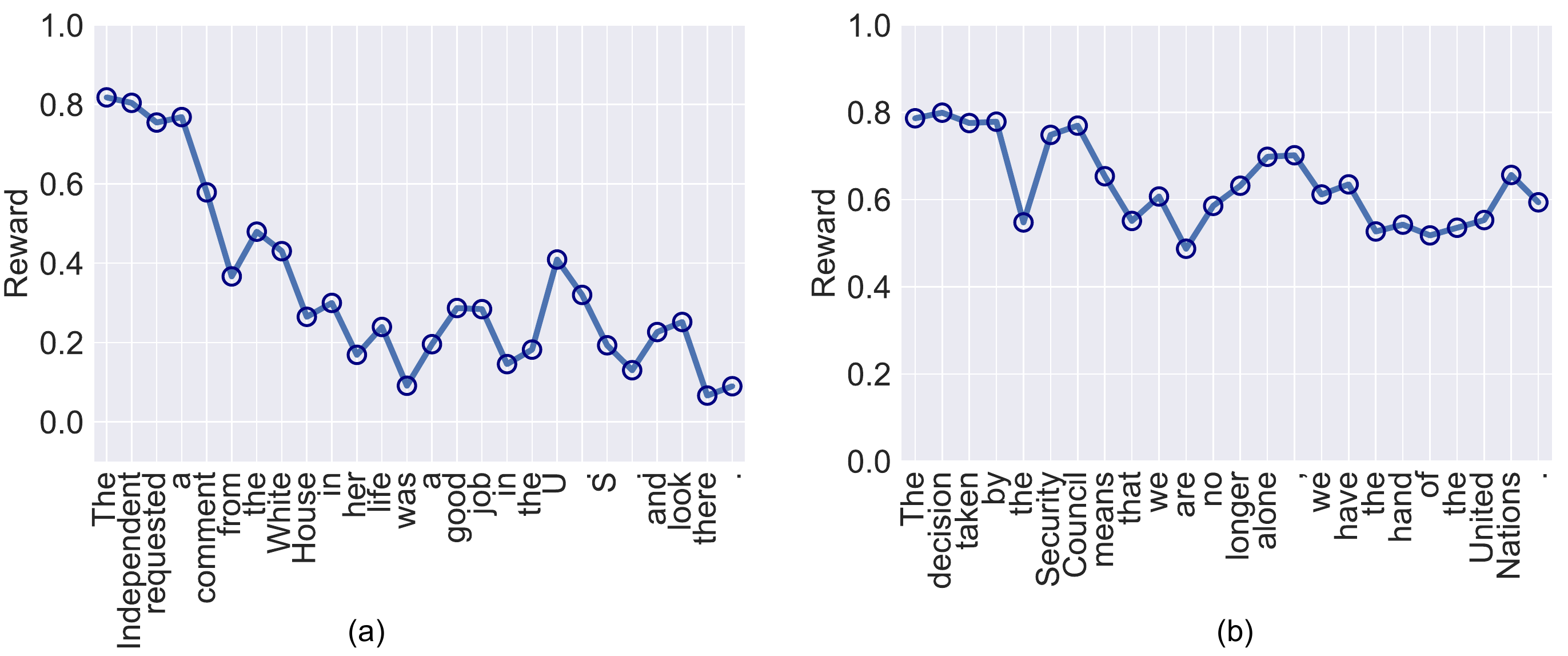} 
	\vspace{-2mm}
	\caption{Guider-Matching Rewards Illustrations.}\label{fig:rewards_ill}
	\vspace{-2mm}
\end{figure}

\paragraph{Case Study of Guider-Matching Rewards}
Figure \ref{fig:rewards_ill}(a) illustrates the feature-matching rewards in the generation. Figure \ref{fig:rewards_ill}(a) shows an example of failure generation in the training stage, when two sentences are combined by the word `\emph{was}'. It is grammatically wrong to select `\emph{was}' for the generator, thus the guider network gives a small reward. We can see that the rewards become lower with more time steps, which is consistent with the exposure bias. Figure \ref{fig:rewards_ill}(b) shows a successful generation, where the rewards given by the guider are relatively high (larger than 0.5). These observations validate that: (\RN{1}) exposure bias exists in MLE training. (\RN{2}) RL training with exploration can help reduce the effects of exposure bias. (\RN{3}) Our proposed feature-matching rewards can provide meaningful guidance to maintain sentence structure and fluency.
\subsection{Non-parallel Text-style Transfer}
We test the proposed framework on the non-parallel text-style-transfer task, where the goal is to transfer one sentence in one style ({\it e.g.}, positive) to a similar sentence but with a different style ({\it e.g.}, negative). 
Pair-wise information should be inferred from the training data, which becomes more challenging. 
For a fair comparison, we use the same data and its split method as in~ \cite{shen2017style}. 
Specifically, there are 444,000, 63,500, and 127,000 sentences with either
positive or negative sentiments in the training, validation and test sets, respectively.

\begin{table}[t!]
	\centering
\begin{adjustbox}{width=\linewidth, tabular=lccc,center}
		\midrule[1.2pt]
		{\bf Model}  &{\bf Acc(\%)}  &{\bf BLEU} &{\bf BLEU-ref}
		\\ 
		\midrule
		CVAE~\cite{shen2017style}    &73.9 &20.7 &7.8 \\ 
		Controllable~\cite{hu2017controllable} &86.7 &58.4 & -  \\
		BackTrans~\cite{prabhumoye2018style} &91.2 &2.8 &2.0\\ 
		DeleteAndRetrieval~\cite{li2018delete} &88.9 &36.8 &14.7 \\ 
		\midrule
		Guider (Ours) & \textbf{92.7} & \textbf{52.1} & \textbf{25.4} \\
		\bottomrule[1.2pt]
	\end{adjustbox}
	\vspace{-2mm}
	\caption{Non-parallel text style transfer results on the test set with \textit{human} references.}
	\label{tab: style_transfer}
\end{table}

To measure whether the original sentences (in the test set) have been transferred to the desired sentiment, we follow the settings of \cite{shen2017style} and employ a pretrained CNN classifier, which achieves an accuracy of $97.4\%$ on the validation set, to evaluate the transferred sentences. We also report the BLEU scores with original sentences (BLEU) and human references (BLEU-ref)~\cite{li2018delete}, to evaluate the content preservation of transferred sentences. Results are summarized in Table~\ref{tab: style_transfer}. 
Our proposed model exhibits higher transfer accuracy and better content preservation, indicating the guider network provides good sentiment guidance to better preserve the content information.
%

\begin{table}[ht]
	\begin{tcolorbox}
		\vspace{-2mm}
		\begin{adjustbox}{width=1.13\linewidth,tabular=l,center}
			\textbf{From positive to negative}\\
			Original: all the employees are \textbf{friendly} and \textbf{helpful} .  \\
			Transferred: all the employees are \textbf{rude} and \textbf{unfriendly} . 
			\vspace{2mm}\\
			Original: i 'm so \textbf{lucky} to have found this place ! \\
			Transferred: i 'm so \textbf{embarrassed} that i picked this place . \\
			\vspace{-2mm}
			\\
			\bottomrule[1.2pt]
			\vspace{1mm}
			\textbf{From negative to positive}\\
			Original: the service was \textbf{slow} . \\
			Transferred: the service was \textbf{fast} and \textbf{friendly} .\vspace{2mm}\\
			Original:  i would \textbf{never} eat there again and would probably \textbf{not stay} there either . \\
			Transferred:  i would \textbf{definitely} eat this place and i would \textbf{recommend} them . \\
		\end{adjustbox}
		\vspace{-2mm}
	\end{tcolorbox}
	\vspace{-1mm}
	\caption{  
		Generated samples of guided style transfer.
	}
	\label{tab:transfer}
\end{table}

\section{Conclusions}

We have proposed a model-based imitation-learning framework for adversarial text generation, by introducing a guider network to model the generation environment.
The guider network provides a plan-ahead mechanism for next-word selection. Furthermore, this framework can alleviate the sparse-reward issue, as the intermediate rewards are used to optimize the generator. Our proposed models are validated on both unconditional and conditional text generation, including adversarial text generation and non-parallel style transfer. We achieve improved performance in terms of generation quality and diversity for unconditional  and conditional generation tasks.

\paragraph{Acknowledgement}
The authors would like to thank the anonymous reviewers for their insightful comments. The research was supported in part by DARPA, DOE, NIH, NSF and ONR.

%
\bibliographystyle{acl_natbib}
\bibliography{reference}

\clearpage

\appendix
\section{Additional Experiments}
\paragraph{More Generated Samples of Text Generation}
Table \ref{tab: wmt_appendix} lists more generated samples on the proposed GMGAN and its baselines. From the experiments, we can see, (i) SeqGAN tends to generate shorter sentences, and the readability and fluency is very poor. (ii) LeakGAN tends to generate very long sentences, and usually longer than the original sentences. However, even with good locality fluency, its sentences usually are not semantically consistent. By contrast, our proposed GMGAN can generate sentences with similar length to the original sentences, and has good readability and fluency. This is also validated in the Human evaluation experiment.
\vspace{-0mm}
\paragraph{Image Captioning} 
We conduct experiments on image captioning~\cite{karpathy2015deep}, investigating benefits brought by the Guider network. In image captioning, instead of using a discriminator to define final rewards for generated sentence, we adopt evaluation metrics computed based on human references. 
The final rewards appear more important as they contain reference (ground-truth) information. Feature-matching rewards work as a regularizer of the final rewards. We call our model in this setting a guider-matching sequence training (GMST) model. An overview of GMST is provided in the Appendix.
We test our proposed model on the MS COCO dataset \cite{karpathy2015deep}, containing 123,287 images in total. Each image is annotated with at least 5 captions. Following Karpathy’s split \cite{karpathy2015deep}, 5,000 images are used for both validation and testing. We report BLEU-$k$ ($k$ from 1 to 4), CIDEr~\cite{vedantam2015cider}, and METEOR~\cite{banerjee2005meteor} scores. We consider two settings: (\emph{i}) using a pre-trained 152-layer ResNet \cite{he2016deep} for feature extraction, where we take the output of the 2048-way \textit{pool5} layer from ResNet-152, pretrained on the ImageNet dataset; and
(\emph{ii}) using semantic tags detected from the image as features~\cite{gan2017semantic}. 
We use an LSTM with 512 hidden units with mini-batches of size 64. Adam \cite{kingma2014adam} is used for optimization, with learning rate $2\times10^{-4}$. 
We pretrain the captioning model for the maximum 20 epochs, then use the reinforcement learning to train it for 20 epochs and test on the best model on the validation set. 

\begin{table}[htp]
	\centering
	\small
	\vspace{2mm}
	\begin{adjustbox}{width=\linewidth,tabular=lcccccc,center}
		\toprule[1.2pt]
		\textbf{Method} & \textbf{BLEU-3} & \textbf{BLEU-4} & \textbf{METEOR} & \textbf{CIDEr} \\
		\midrule
		\multicolumn{3}{c}{\emph{No attention, Greedy, Resnet-152}} \\
		MLE  & 37.2 & 26.5 & 23.1 & 83.9 \\
		Guider~  & 38.0 & 27.3 & 23.9 & 85.4 \\
		MIXER (BLEU)~  & 39.1 & 29.3 & 22.3 & 79.7 \\
		SCST (BLEU)~ & 41.6 & 31.6 & 23.1 & 87.5 \\
		GMST (BLEU)~  & \textbf{41.8}& \textbf{32.1} & 23.4 & 87.9 \\
		MIXER (CIDEr)~  & 39.1 & 27.7 & 23.0 & 90.9 \\
		SCST (CIDEr)~  & 41.2 & 30.0 & 24.3 & 98.6 \\
		GMST (CIDEr)~  & 41.3 & 30.3 & \textbf{24.4} & \textbf{100.1}\\
		\midrule  
		\multicolumn{2}{c}{\emph{No attention, Greedy, Tag}}\\
		MLE &  39.4 & 28.8 & 24.4 & 91.3 \\
		Guider~  & 39.6 & 29.0 & 24.6 & 92.7 \\
		MIXER (BLEU)~  & 42.4 & 32.2 & 23.7 & 90.4 \\
		SCST (BLEU)~  & 43.9 & 33.6 & 24.5 & 95.9 \\
		GMST (BLEU)~  & \textbf{44.3} & \textbf{33.9}& 24.5 & 97.1 \\
		MIXER (CIDEr)~  & 42.1 & 30.8 & 24.7 & 101.2 \\
		SCST (CIDEr)~  & 43.6 & 32.1 & 25.4 & 105.5 \\
		GMST (CIDEr)~  & 44.1 & 32.6 & \textbf{25.5} & \textbf{107.4} \\
		\bottomrule[1.2pt]
	\end{adjustbox}
	\vspace{-2mm}
	\caption{Results for image captioning on the MS COCO dataset; the higher the better for all metrics. 
	} 	\label{table:captioning}
	\vspace{-3mm}
\end{table}
%
%
The results are summarized in Table~\ref{table:captioning}. When comparing an AutoEncoder (AE) with a variant implemented by adding a guider network (Guider), improvements are observed. We compare the proposed GMST with SCST. Note the main difference between GMST and SCST is that the former employs our proposed feature-matching reward, while the latter only considers the final reward provided by evaluation metrics. GMST achieves higher scores compared with SCST on its optimized metrics. The gain of GMST compared with SCST comes from the immediate rewards, which can maintain the semantic consistency and sentence structure, preventing language-fluency damage caused by only focusing on evaluation metrics. Specifically, the average length of generated sentence with a Guider is 15.7, and 12.9 for traditional generator. 

\paragraph{Comparison with MLE} The guider network models the long-term dependency and overcome the issue of sparse reward inspired by model predictive control (MPC). The experiments aim to quantify the gain when incorporating MPC for imitation learning, i.e., MLE and RL finetune. 

We provide an additional comparison with \citet{caccia2018language} and evaluate the diversity and quality with BLEU scores. We also report the F1-BLEU which considers both diversity and quality:

\begin{table*}[t]
	\centering
\begin{adjustbox}{width=\linewidth,tabular=l r  c  c  r c c r c c , center}
\toprule[1.2pt]
\textbf{Method} & \textbf{Test-BLEU-2} & \textbf{3} & \textbf{4} & \textbf{Self-BLEU-2} & \textbf{3} & \textbf{4} & \textbf{F1-BLEU-2} & \textbf{3} & \textbf{4}\\
\midrule[1.2pt]
MLE~\cite{caccia2018language} & 0.902 &  0.706 &  0.470 &  0.787 & 0.646 & 0.485 & \textbf{0.345} & 0.472 & 0.491\\

Guider (MLE)	&	      0.920 &  0.723 & 0.489  &    0.812 & 0.589 & 0.360 &	0.312&	0.524&	0.554\\

GMGAN (Ours) & 0.923 &  0.727 & 0.491 & 0.814 & 0.576 & 0.328 &	0.310&	\textbf{0.537}&	\textbf{0.567}\\
\bottomrule[1.2pt]
\end{adjustbox}
\vspace{-2mm}
\caption{Additional Comparison with MLE~\cite{caccia2018language} .}
\vspace{-2mm}
\end{table*}
\section{Discussions of the Guider Network}
Guider network can be regarded as a model of the text-generation environments, namely  
the model of dynamics. It takes current $\sv_t$ and $\av_t$ as input, and outputing an estimate of the next state $\sv_{t+\triangle t}$ at time $t+\triangle t$. In the text generation setting, when $\triangle t=1$, we can exactly get the feature representation of the current generated sentence if the guider does not help the word selection. If not, we cannot exactly get this feature extraction since the guider's prediction partly determine next token.
In practice, we use $\triangle t=c=4$, to give the guider planning ability, to help for word selection and guide sentence generation.

\section{Experimental Setup}
\subsection{Adversarial Text Generation}
For Image COCO, the learning rate of the generator is 0.0002, the learning rate  of the guider 0.0002, the maximum length of sequence is 25.
For WMT, the learning rate  of the guider 0.0002, the learning rate  of the guider 0.0002, the maximum length of sequence is 50. 
We use $c=4$ chosen from $[2,3,4,5,8]$ and $\gamma=0.25$ chosen from $[0.1, 0.25, 0.5, 0.75, 0.99]$. 
We use Adam~\cite{kingma2014adam} optimization algorithm to train the guider, generator and discriminator.

For both tasks, the LSTM state of dimension for the generator is 300, and the LSTM state of dimension for the generator is 300. 
The dimension of word-embedding is 300. The output dimension of the linear transformation connecting guider and generator is 600$\times$10. The learning rate of Discriminator is 0.001.

\subsection{Conditional Generation}
For Image Captioning, the learning rate  of the guider 0.0002, the learning rate  of the guider 0.0002, the maximum length of sequence is 25. For Style transfer, the learning rate  of the guider 0.0001, the learning rate  of the guider 0.0001, the maximum length of sequence is 15. 

\subsection{Network Structure of Models}
\begin{table}[h!]
	\small
	\centering
	\begin{tabular}{c}
		\toprule[1.2pt]
		(Sub-)sequence to latent features \\
		\midrule
		Input 300$\times$ Seq. Length Sequences\\
		
		\midrule
			$5\times 300$ conv. 300 ReLU, stride 2\\
			$5\times 1$  conv. 600 ReLU, stride 2\\
		MLP output 600, ReLU\\
		\bottomrule[1.2pt]
	\end{tabular} 
\vspace{-2mm}
\caption{Architecture of Encoder.}
	\label{Table:TriGAN model_celebA}
\end{table}

The LSTM state of dimension for the generator is 300, and the LSTM state of dimension for the guider is 300. The dimension of word-embedding is 300. 

\begin{table}[h!]
	\small
	\centering
	\begin{tabular}{c}
		\toprule[1.2pt]
		Sequence to a scalar value \\
		\midrule
		Input 300$\times$ Seq. Length Sequences\\
		
		\midrule
		$5\times 300$ conv. 300 ReLU, stride 2\\
		$5\times 1$  conv. 600 ReLU, stride 2\\
		MLP output 1, ReLU\\
		\bottomrule[1.2pt]
	\end{tabular} 
\vspace{-2mm}
\caption{Architecture of Discriminator.}
	\label{Table:TriGAN model_celebA}
\end{table}

\newpage

\section{Algorithm Details}
\begin{algorithm}
	\caption{Guider Matching Generative Adversarial Network (GMGAN)}
	\label{algo:algo1}
	\begin{algorithmic}[1]
		\REQUIRE generator policy $\pi^{\phi}$; discriminator $D_\theta$; guider network $G^\psi$; a sequence dataset $\mathcal{S}=\{X_{1 \ldots T}\}$.
		\STATE Initialize $G^\psi$, $\pi^{\phi}$, $D^\theta$ with random weights.
		\STATE Pretrain generator $\pi^{\phi}$, guider $G^\psi$ and discriminator $D^\theta$ with MLE loss.
		\REPEAT 
		\FOR {g-steps}
		\STATE{Generate a sequence $Y_{1\ldots T} \sim \pi^{\phi}$.}
		\STATE {Compute $Q_t$ via (5), and update $\pi^\phi$ with policy gradient via (8).}
		\ENDFOR
		\FOR {d-steps}
		\STATE {Generate a sequences from $\pi^{\phi}$.}
		\STATE {Train discriminator $D_\theta$.}
		\ENDFOR
		\UNTIL{GMGAN converges}
	\end{algorithmic}
\end{algorithm}

\begin{figure*}[t!] 
	\centering
	\includegraphics[width=1.0\linewidth]{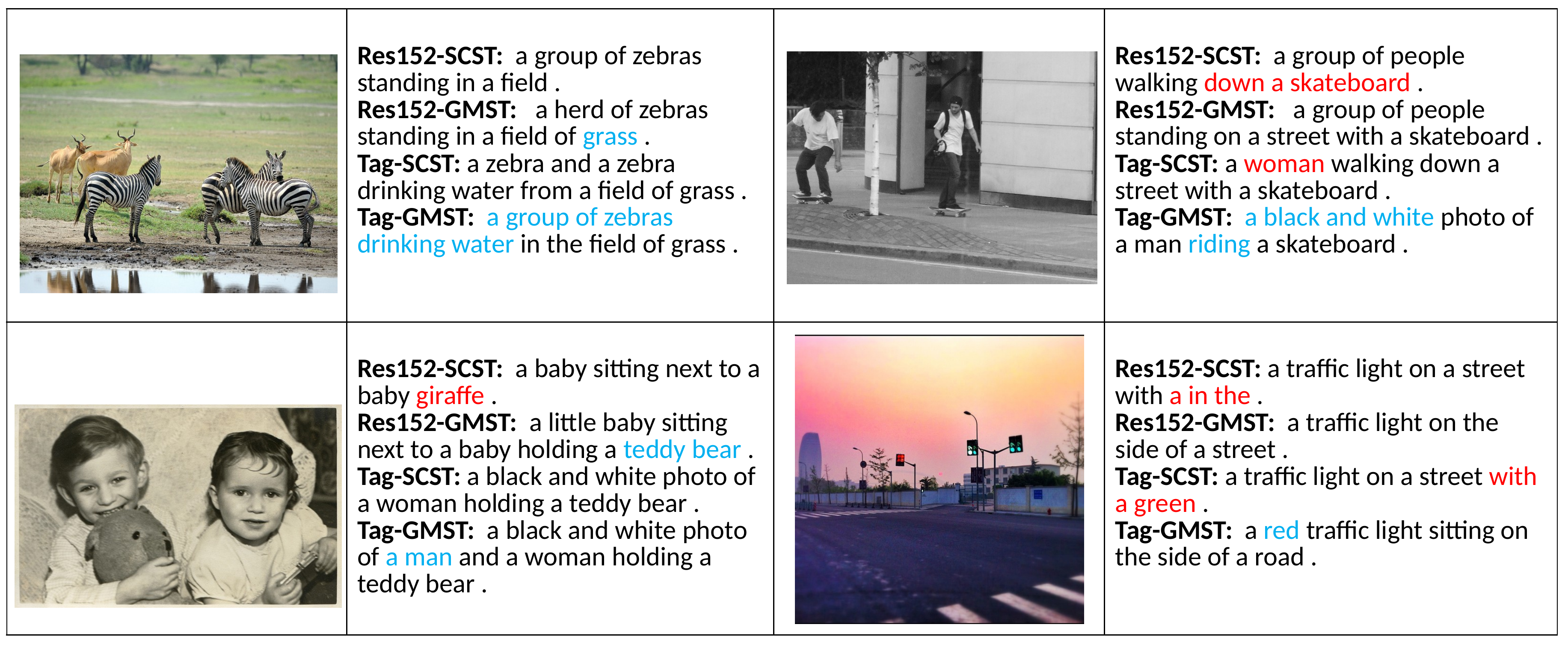}  
	\caption{Examples of image captioning on MS COCO.}
	\label{fig:COCO}
	\vspace{-5mm}
\end{figure*}

\begin{algorithm}
	\caption{Guider Matching Sequence Training (GMST)}
	\label{algo:algo1}
	\begin{algorithmic}[1]
		\REQUIRE generator policy $\pi^{\phi}$; discriminator $D_\theta$; guider network $G^\psi$; a sequence dataset $\mathcal{S}=\{Y_{1 \ldots T}\}$ and its condition information $\mathcal{I}=\{X\}$
		\STATE Initialize $G^\psi$, $\pi^{\phi}$, $D^\theta$ with random weights.
		\STATE Pretrain generator $\pi^{\phi}$, guider $G^\psi$ and discriminator $D^\theta$ with MLE loss.
		\REPEAT 
		\STATE{Generate a sequence $Y_{1\ldots T} \sim \pi^{\phi}$.}
		\STATE{Compute evaluation scores based on references.}
		\STATE {Compute $Q_t^s$ via (6), and update $\pi^\phi$ with policy gradient via (8).}
		\UNTIL{GMST converges}
	\end{algorithmic}
\end{algorithm}

\newpage

\begin{table*}[t!]
	\centering
	\begin{tcolorbox}
	\begin{adjustbox}{width=\textwidth,tabular=l  p{21.9cm},center}
		\textbf{Method} & \textbf{Generated Examples}\\
		\hline
		\textbf{Real Data} &
		What this group does is to take down various different websites it believes to be criminal and leading to terrorist acts .\newline
		Over 1 , 600 a day have reached Greece this month , a higher rate than last July when the crisis was already in full swing .\newline
		" We ' re working through a legacy period , with legacy products that are 10 or 20 years old ," he says .\newline
		' The first time anyone says you need help , I ' m on the defensive , but that ' s all that I know .\newline
		Out of those who came last year , 69 per cent were men , 18 per cent were children and just 13 per cent were women .\newline
		He has not played for Tottenham ' s first team since and it is now nearly two years since he completed a full Premier League match for the club .\newline
		So you have this man who seems to represent this way to live and how to be a good citizen of the world .\newline
		CNN : You made that promise , but it wasn ' t until 45 years later that you acted on it .\newline
		This is a part of the population that is notorious for its lack of interest in actually showing up when the political process takes place .\newline
		They picked him off three times and kept him out of the end zone in a 22 - 6 victory at Arizona in 2013 .\newline
		The treatment was going to cost £ 12 , 000 , but it was worth it for the chance to be a mum .\newline
		But if black political power is so important , why hasn ' t it made more of a difference in the lives of poor black people in Baltimore such as Gray ?\newline
		Local media reported the group were not looking to hurt anybody , but they would not rule out violence if police tried to remove them .\newline
		The idea was that couples got six months ' leave per child with each parent entitled to half the days each .\newline
		The 55 to 43 vote was largely split down party lines and fell short of the 60 votes needed for the bill to advance .\newline
		Taiwan ' s Defence Ministry said it was " aware of the information ," and declined further immediate comment , Reuters reported .\newline
		I ' m racing against a guy who I lost a medal to - but am I ever going to get that medal back ?\newline
		Others pushed back their trips , meaning flights early this week are likely to be even more packed than usual .\newline
		" In theory there ' s a lot to like ," Clinton said , " but ' in theory ' isn ' t enough .\newline
		If he makes it to the next election he ' ll lose , but the other three would have lost just as much .
		\\
		\hline
		\textbf{SeqGAN} & Following the few other research and asked for " based on the store to protect older , nor this . \newline
		But there , nor believe that it has reached a the person to know what never - he needed . \newline
		The trump administration later felt the alarm was a their doctors are given . \newline
		We have been the time of single things what people do not need to get careful with too hurt after wells then . \newline
		If he was waited same out the group of fewer friends a more injured work under it . \newline
		It will access like the going on an " go back there and believe . \newline
		Premier as well as color looking to put back on a his is . \newline
		So , even though : " don ' t want to understand it at an opportunity for our work . \newline
		I was shocked , nor don ' t know if mate , don ' t have survived ,  \newline
		So one point like ten years old , but a sure , nor with myself more people substantial . \newline
		And if an way of shoes of crimes the processes need to run the billionaire . \newline
		Now that their people had trained and people the children live an actor , nor what trump had . \newline
		However , heavily she been told at about four during an innocent person . \\
		\hline
		\textbf{LeakGAN} & The country has a reputation for cheap medical costs and high - attack on a oil for more than to higher its - wage increase to increase access to the UK the UK women from the UK ' s third nuclear in the last couple of weeks .\newline
		I ' ve been watching it through , and when the most important time it is going to be so important .\newline
		I ' m hopeful that as that process moves along , that the U . S . Attorney will share as much as far as possible .\newline
		The main thing for should go in with the new contract , so the rest of the Premier League is there to grow up and be there ," she said .\newline
		I think the main reason for their sudden is however , I didn ' t get any big thing ," he says , who is the whole problem on the U . S . Supreme Court and rule had any broken .\newline
		The average age of Saudi citizens is still very potential for the next year in the past year , over the last year he realised he has had his massive and family and home .\newline
		" I think Ted is under a lot of people really want a " and then the opportunity to put on life for security for them to try and keep up .\newline
		The new website , set to launch March 1 , but the U . S is to give up the time the case can lead to a more than three months of three months to be new home .\newline
		It ' s a pub ; though it was going to be that , but , not , but I am not the right thing to live ," she said .\newline
		" I ' m not saying method writing is the only way to get in the bedroom to get through the season and we ' ll be over again ," he says .\newline
		I ' m not suggesting that our jobs or our love our years because I have a couple of games where I want it to be .\newline
		The German government said 31 suspects were briefly detained for questioning after the New Year ' s Eve trouble , among them not allowed to stay in the long - term .\newline
		It was a punishment carried out by experts in violence , and it was hard to me he loved the man and he ' s got off to support me in the future .\newline
		" I ' ve known him , all that just over the last two weeks and for the last 10 years , I ' ll have one day of my life ," she said .\newline
		The main idea behind my health and I think we saw in work of our country was in big fourth - up come up with a little you ' ve ever .\newline
		he Kings had needed scoring from the left side , too , and King has provided that since his return are the of the first three quarters of the game .\newline
		The average number of monthly passengers arriving at the University of January 1 . 1 million people and another average visit men were on the year .\newline
		It ' s going to be a good test for us and we are on the right way to be able to get through it on every day on the year .\\
		\hline
		\textbf{GMGAN} & 
		But it ' s grown up a little now , and might be ready for actually putting into your house . \newline
More than a dozen Republicans and a handful of Democrats have announced they are running for their party ' s 2016 presidential nomination , and when they were wealthy in 2010 right , what he has . \newline
And with a growing following of more than 45 , 000 people on Facebook , awareness of their work is on the rise .\newline
In all age groups , for instance , more people cited retirement as the reason for being out of the labour force , and it wasn ' t a problem in big .\newline
I had to train really , really hard and that ' s the advice I can give , because if you don ' t work hard somebody else will .\newline
I am picking up two cars tomorrow and taking them down south tomorrow if all goes according to plan ," he said .\newline
The team looked into the influence of marriage on weight loss after surgery - as well as the effects of surgery on the quality of his administration and rest on the world .\newline
Two former prime ministers were set to face off in the second round of a presidential election in New Hampshire .\newline
A third more complaints were made about the accounts between April and December last year than in the whole of 2014 / 15 .\newline
United Airlines subsequently worked to get those passengers back in the air so they could get to Colorado , the airline spokesman said .\newline
Mr Brown was standing in the kitchen when he started to feel a bit cold - and he noticed the door had disappeared .\newline
She has focused instead on where she parts ways with her rival on other issues , like to have someone with a president has revealed .\newline
Once , an ex - boyfriend and I lived with her for two months after we came back from travelling .\newline
He had faced 10 years in prison on the charges but the first government have been made at the recent peak .\newline
" We weren ' t exposed to things we didn ' t have in the same way kids these days are ," said Obama .\newline
I have no idea what it is , but there is definitely an intelligence - a higher intelligence - at work you have you want to make sure you are going into the local community .\newline
His current club have confirmed they would be willing to listen to offers for the attacking midfielder , but we did not have the right manager - there ' s summer to be in a big .\newline
We are in the last 16 and the target is always to win in the Champions League and will continue at the best level to be the coach .\newline
People are seeing that you can go into real estate and do really well and do something we want and if we make the right decision , and how we will be doing it is .\\
	\end{adjustbox}
	\end{tcolorbox}
	\caption{
		Generated Examples on EMNLP2017 WMT.
	}
	\label{tab: wmt_appendix}
\end{table*}
\newpage

\begin{table*}[h]
	\centering
	\begin{tcolorbox}
	\begin{adjustbox}{scale=0.9,tabular=l  p{13.4cm},center}
				Original: & i 'm so lucky to have found this place ! \\
				Guider: &	i 'm so embarrassed that i picked this place . \\
				\hline
				Original: & awesome place , very friendly staff and the food is great !  \\
				Guider: & disgusting place , horrible staff and extremely rude customer service . \\
				\hline
				Original: & this was my first time trying thai food and the waitress was amazing ! \\
				Guider: & this was my first experience with the restaurant and we were absolutely disappointed .  \\
				\hline
				Original: & thanks to this place ! \\
				Guider: & sorry but this place is horrible . \\
				\hline
				Original: & the staff was warm and friendly . \\
				Guider: & the staff was slow and rude .  \\
				\hline
				Original: & great place and huge store . \\
				Guider: & horrible place like ass screw .  \\
				\hline
				Original: & the service is friendly and quick especially if you sit in the bar . \\
				Guider: &	the customer service is like ok - definitely a reason for never go back .. \\
				\hline
				Original: & everything is always delicious and the staff is wonderful . \\
				Guider: & everything is always awful and their service is amazing . \\
				\hline
				Original: & best place to have lunch and or dinner . \\
				Guider: & worst place i have ever eaten . \\
				\hline
				Original: & best restaurant in the world ! \\
				Guider: & worst dining experience ever !  \\
				\hline
				Original: & you 'll be back ! \\
				Guider: &	you 're very disappointed ! \\
				\hline
				Original: & you will be well cared for here ! \\
				Guider: & you will not be back to spend your money .  \\
				\hline
				Original: & they were delicious ! \\
				Guider: & they were overcooked .  \\
				\hline
				Original: & seriously the best service i 've ever had . \\
				Guider: & seriously the worst service i 've ever experienced . \\
				\hline
				Original: & it 's delicious ! \\
				Guider: &	it 's awful . \\
	\end{adjustbox}
	\end{tcolorbox}
	\label{tab: trans_samples1}
	\caption{Sentiment transfer samples on Yelp dataset (positive $\to$ negative).}
\end{table*}

\begin{table*}[h]
	\centering
	\begin{tcolorbox}
	\begin{adjustbox}{scale=0.9,tabular=l  p{13.4cm},center}
		Original: & gross !  \\
		Guider: & amazing !  \\
		\hline
		Original: & the place is worn out . \\
		Guider: &	the place is wonderful . \\
		\hline
		Original: & very bland taste .  \\
		Guider: & very fresh . \\
		\hline
		Original: & terrible service ! \\
		Guider: & great customer service !  \\
		\hline
		Original: & this place totally sucks . \\
		Guider: & this place is phenomenal .  \\
		\hline
		Original: & this was bad experience from the start . \\
		Guider: & the food here was amazing good .  \\
		\hline
		Original: & very rude lady for testing my integrity . \\
		Guider: & very nice atmosphere for an amazing lunch ! \\
		\hline
		Original: & they recently renovated rooms but should have renovated management and staff . \\
		Guider: & great management and the staff is friendly and helpful .  \\
		\hline
		Original: & this store is not a good example of sprint customer service though . \\
		Guider: &	this store is always good , consistent and they 're friendly . \\
		\hline
		Original: & one of my least favorite ross locations . \\
		Guider: & one of my favorite spots .  \\
		\hline
		Original: & horrible in attentive staff . \\
		Guider: & great front desk staff !  \\
		\hline
		Original: & the dining area looked like a hotel meeting room . \\
		Guider: & the dining area is nice and cool .  \\
		\hline
		Original: & never ever try to sell your car at co part ! \\
		Guider: & highly recommend to everyone and recommend this spot for me ! \\
		\hline
		Original: & i ordered the filet mignon and it was not impressive at all . \\
		Guider: &	i had the lamb and it was so good .\\
	\end{adjustbox}
	\end{tcolorbox}
	\label{tab: trans_samples}
	\caption{Sentiment transfer samples on Yelp dataset (negative $\to$ positive).}
\end{table*}

\end{document}